%% file: main.tex
\documentclass[runningheads]{llncs}
\usepackage[T1]{fontenc}
\usepackage{amsmath}
\usepackage{amssymb}
\usepackage{booktabs}

\usepackage{graphicx,verbatim}
\usepackage{xcolor}
\usepackage{bm}
\usepackage{colortbl}
\usepackage{multirow}
\usepackage{pifont}
\usepackage{amsfonts}

\usepackage[capitalize]{cleveref}
\crefname{section}{Sec.}{Secs.}
\Crefname{section}{Section}{Sections}
\Crefname{table}{Table}{Tables}
\crefname{table}{Tab.}{Tabs.}
\begin{document}

\title{Iterative Deployment Exposure for Unsupervised Out-of-Distribution Detection}
\titlerunning{Iterative Deployment Exposure}
%
\author{Lars Doorenbos\inst{1,2}\orcidID{0000-0002-0231-9950} \and
Raphael Sznitman\inst{1}\orcidID{0000-0001-6791-4753} \and
Pablo Márquez-Neila\inst{1}\orcidID{0000-0001-5722-7618}}
\authorrunning{L. Doorenbos et al.}
%
\institute{University of Bern, Switzerland \and
University of Bonn, Germany\\
\email{doorenbos@iai.uni-bonn.de}}



\maketitle              
\begin{abstract}
Deep learning models are vulnerable to performance degradation when encountering out-of-distribution (OOD) images, potentially leading to misdiagnoses and compromised patient care. These shortcomings have led to great interest in the field of OOD detection. 
Existing unsupervised OOD (U-OOD) detection methods typically assume that OOD samples originate from an unconcentrated distribution complementary to the training distribution, neglecting the reality that deployed models passively accumulate task-specific OOD samples over time.
To better reflect this real-world scenario, we introduce Iterative Deployment Exposure (IDE), a novel and more realistic setting for U-OOD detection.
We propose CSO, a method for IDE that starts from a U-OOD detector that is agnostic to the OOD distribution and slowly refines it during deployment using observed unlabeled data. 
CSO uses a new U-OOD scoring function that combines the Mahalanobis distance with a nearest-neighbor approach, along with a novel confidence-scaled few-shot OOD detector to effectively learn from limited OOD examples. We validate our approach on a dedicated benchmark, showing that our method greatly improves upon strong baselines on three medical imaging modalities.

\keywords{Out-of-Distribution Detection \and Deployment \and Reliability.}

\end{abstract}
%
%
%

\input{defs}

\input{01introduction}
\input{03method}
\input{04experiments}

\input{05discussion}

\input{06conclusion}

%
%
%
\bibliographystyle{splncs04}
\bibliography{bib}




\end{document}

%% file: defs.tex

\newif\ifdraft
\drafttrue

\definecolor{orange}{rgb}{1,0.5,0}
\definecolor{gr}{rgb}{0,0.65,0}
\definecolor{mygray}{gray}{0.95}

\ifdraft
 \newcommand{\RS}[1]{{\color{red}{\bf RS: #1}}}
 \newcommand{\rs}[1]{{\color{red}#1}}
 \newcommand{\PMN}[1]{{\color{orange}{\bf PMN: #1}}}
 \newcommand{\pmn}[1]{{\color{orange}#1}}
 \newcommand{\LD}[1]{{\color{blue}{\bf LD: #1}}}
 \newcommand{\ld}[1]{{\color{blue}#1}}
 \newcommand{\old}[1]{{\color{gr}#1}}
\else
 \renewcommand{\sout}[1]{}
 \newcommand{\RS}[1]{{\color{red}{}}}
 \newcommand{\rs}[1]{#1}
 \newcommand{\PMN}[1]{{\color{red}{}}}
 \newcommand{\pmn}[1]{#1}
\fi

\newcommand{\real}{\mathbb{R}}
\newcommand{\x}{\mathbf{x}}
\newcommand{\z}{\mathbf{z}}
\newcommand{\y}{\mathbf{y}}
\newcommand{\haty}{\hat{\y}}
\newcommand{\w}{\mathbf{w}}
\renewcommand{\d}{\mathbf{d}}
\newcommand{\D}{\mathcal{D}}
\newcommand{\X}{\mathcal{X}}
\newcommand{\Z}{\mathcal{Z}}
\newcommand{\J}{\mathbf{J}}
\newcommand{\bZ}{\mathbf{Z}}
\newcommand{\M}{\mathcal{M}}
\newcommand{\I}{\mathcal{I}}
\newcommand{\bmu}{\bm{\mu}}
\newcommand{\covar}{\mathbf{\Sigma}}
\newcommand{\jacobian}{\mathbf{J}}
\newcommand{\balpha}{\bm{\alpha}}
\newcommand{\pkde}{p_{\textrm{kde}}}
\newcommand{\psv}{p_{\balpha}}
\newcommand{\f}{\mathbf{f}}
\newcommand{\g}{\mathbf{g}}
\newcommand{\F}{\mathcal{F}}
\renewcommand{\a}{\mathbf{a}}
\newcommand*{\XX}{%
  \textsf{X\kern-1ex X}%
}
\newcommand{\pin}{\mathcal{P}_{\textrm{in}}}
\newcommand{\pout}{\mathcal{P}_{\textrm{out}}}
\newcommand{\pdeploy}{\mathcal{P}_{\textrm{deploy}}}
\newcommand{\sigmain}{\sigma^\textrm{in}}
\newcommand{\sigmaout}{\sigma^\textrm{out}}

\newcommand{\xmark}{\ding{55}}%
\newcommand{\cmark}{\ding{51}}%

\newcommand{\oc}{{\bf{One-class}}}
\newcommand{\mc}{{\bf{Multi-class}}}
\newcommand{\hr}{{\bf{High-resolution}}}

\newcommand{\overbar}[1]{\mkern 1.5mu\overline{\mkern-1.5mu#1\mkern-1.5mu}\mkern 1.5mu}

\renewcommand{\vec}[1]{\mathbf{#1}}

%% file: 01introduction.tex
\section{Introduction}

Deep learning (DL) models fundamentally rely on the premise that the training data distribution aligns with that of the test data. However, this assumption often fails in real-world situations where the performance of a DL model diverges from its initial benchmark due to encounters with out-of-distribution (OOD) samples. This phenomenon is highly problematic in medical imaging, where the dependability of DL models is critical for safe, prolonged use in the field.


\begin{figure*}[t]
    \centering
    \includegraphics[width=0.95\linewidth]{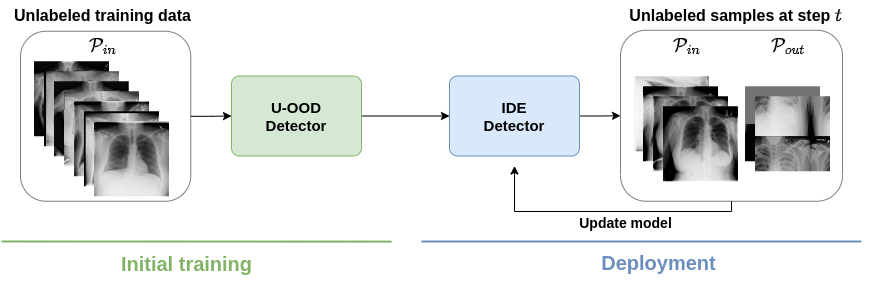}
    \caption{\textbf{Iterative deployment exposure.} The initial detector, trained on unlabeled ID data, is iteratively refined with unlabeled deployment samples.}
    \label{fig:teaser}
\end{figure*}

These shortcomings have sparked great interest in the field of OOD detection~\cite{hendrycks2016baseline,zhang2021out,zimmerer2022mood}, and unsupervised OOD (U-OOD) detection in particular. Unlike supervised OOD detection, U-OOD assumes neither access to training labels nor OOD samples, thereby encompassing a more generally applicable albeit more challenging setting~\cite{doorenbos2022data,gonzalez2022distance,graham2015kaggle,linmans2024diffusion,marquez2019image,naval2021implicit}. 
The core principle of U-OOD is to identify the level sets of the in-distribution (ID) training data and establish a threshold to distinguish OOD samples, where the assumption is that OOD samples are the complement of the ID and therefore unconcentrated~\cite{scholkopf2002learning,steinwart2005classification}. 


However, this assumption fails to consider the underlying motivation of OOD detection, that is, improving the reliability of \textit{deployed} downstream models.
Treating OOD data as arising from a generic, unconcentrated distribution is a disconnect from the reality that deployment environments inherently impose a specific, concentrated OOD context. 
We argue that the OOD context must be inferred from the deployment setting, and, in turn, the deployment context fully determines a \textit{concentrated} OOD distribution. As this context depends on the deployment, U-OOD works that synthesize anomalies \textit{a priori} (e.g.,~\cite{schluter2022natural,tan2021detecting}) to simulate the OOD distribution are inadequate. Instead, an approach where an initial U-OOD detector, agnostic to the OOD distribution, is gradually updated during deployment to consider the actual OOD distribution is needed (\cref{fig:teaser}). 


Yet, current U-OOD research has mainly neglected this important consideration. While related fields like OOD detection with in-the-wild data~\cite{du2024does,katz2022training} and OOD test-time adaptation~\cite{fan2022simple,zhang2023model} exist, they are predicated on the availability of a large number of test samples: the former relies on a large set of unlabeled data consisting of both ID and OOD samples, while the latter uses the entire unlabeled test set to update a model. In both cases, hundreds of OOD samples are used to update their respective models, and performance is only measured after seeing all of them. Consequently, this is often unrealistic, considering that OOD samples can be scarce in medical applications. 

Instead, we focus on improving detection with only a few OOD samples and tackle the realistic setting where this process is iterative: detect OOD samples, update the OOD detector, and repeat the procedure with a refined model several times. Consequently, evaluating U-OOD detectors should extend beyond a single instance and consider their effectiveness \textit{over time}. We refer to this setting as \textit{IDE} (Iterative Deployment Exposure). Hence, in this work, we,
\begin{enumerate}
    \item introduce the problem of IDE for U-OOD detection, which closely matches the reality of model deployment. 
    \item introduce new metrics and benchmarks to evaluate methods in the IDE setting.
    \item propose CSO, a novel method for IDE that outperforms strong baselines from related fields.
\end{enumerate}

%% file: 03method.tex
\section{Iterative Deployment Exposure}

In U-OOD detection, the training distribution of the downstream model is called the \emph{in distribution}~$\pin$, and the \emph{out distribution}~$\pout$ is assumed to be its \emph{complement}. Given its large support, producing a set of OOD samples representative of~$\pout$ for supervised classification of ID~vs.~OOD samples is intractable. Instead, U-OOD detection methods rely on in-distribution~(ID) samples to train a detector~$\sigmain \colon \mathcal{X} \to \real$ that scores the \emph{OOD-ness} of test samples at inference time. 
Critically, the likelihood of observing a specific image from ~$\pout$ is not uniform once a downstream task is established and the model is deployed in a given environment. Instead, OOD samples seen during the operation of the deployed system constrain ~$\pout$ to the specific application. The goal of IDE is to progressively enrich the detector~$\sigmain$ with observed OOD samples, thus adapting the detection model to the deployment environment. 

More specifically, an IDE system builds a sequence of detectors~$s_t(\x) \colon \mathcal{X} \mapsto \real$ for time steps~$t\in\{0, 1, 2, \ldots\}$. The sequence starts with the base U-OOD detector, $s_0=\sigmain$, and progressively adapts to the OOD samples observed after deployment. At each time~$t$, the detector~$s_t$ is trained with the dataset of samples observed until time step~$t$, denoted $\D_t=\D^{\textrm{train}} \cup \D_t^{\textrm{deploy}}$. 
We assume that~$\D_0^\textrm{deploy} = \varnothing$. 
Crucially, not all elements in~$\D_t$ are labeled as ID or OOD. While elements of $\D^{\textrm{train}}$ are known to come from distribution~$\pin$ and are, therefore, ID~samples, elements of $\D_t^{\textrm{deploy}}$ are unlabeled. These samples of the \emph{deployment distribution}~$\pdeploy$, following~\cite{du2024does,katz2022training}, are modeled with the Huber contamination model~\cite{huber1992robust},
\begin{equation}
    \pdeploy = (1 - \pi)\pin + \pi \pout,
\end{equation}
with contamination ratio~$\pi$. 

To address the lack of labels in $\D_t^{\textrm{deploy}}$,~$s_{t-1}$ is used to pseudo-label the elements of~$\D_t^\textrm{deploy}$ for training~$s_t$. Hence, at each time~$t$,~$\D_t$ can be split into two disjoint subsets~$\D_t^\textrm{in}$ and~$\D_t^\textrm{out}$ containing the samples labeled as ID and OOD, respectively. 
For simplicity and without loss of generality, we assume that the deployment dataset~$\D_t^{\textrm{deploy}}$ grows $K$~elements at each time step, whereby $\left|\D_t^{\textrm{deploy}}\right| = t\cdot{}K$. We will also omit the index~$t$ where not explicitly needed.

\subsection{Model}

The main challenge of OOD~detection is the scarcity of representative OOD~samples. In IDE, the expected number of OOD~samples at time step~$t$ is $\pi\cdot{}t\cdot{}K$, which for a small~$t$ is too small to effectively train a binary classifier, preventing us from simply using binary classifiers to model the detector~$s_t$. As~$t$ increases, however, the feasibility of training a binary classifier improves. We, therefore, design our detection model~$s_t$ to behave as a few-shot learner~$s^-$ when $t$ is close to~$0$ and to gradually transition towards a strong binary learner~$s^+$ as $t$~increases. Formally, we model the detector~$s_t$ as a convex combination of two learner modalities controlled by a mixing factor~$\alpha_t$,
\begin{equation}
s_t(\x) = (1-\alpha_t) s_t^-(\x) + \alpha_t s_t^+(\x).
\end{equation}
The key difference between the learners~$s^-$ and~$s^+$ lies in their inductive biases. The few-shot learner~$s^-$ is a low-variance/high-bias classifier with strong assumptions about in- and out-distributions. Its design is based on the U-OOD detector~$\sigmain$, as detailed below. In contrast, the strong learner~$s^+$ is a low-bias binary classifier, and its architecture can be chosen according to the nature of the input space~$\mathcal{X}$ (e.g.,~a CNN or a transformer architecture for image data), as our approach is agnostic to the internal specifics of~$s^+$. At each step~$t$, both $s^-_t$ and $s^+_t$ are trained independently with the dataset~$\D_t$ pseudo-labeled with~$s_{t-1}$.

The factor $\alpha_t\in[0, 1]$ controls the transition between both models and is proportional to the number of elements pseudo-labeled as OOD,
\begin{equation}
    \alpha_t = \min\left(1, \beta\cdot\left|\D_t^\textrm{out}\right|\right),
\end{equation}
where the factor~$\beta$ is a hyperparameter of our method. We refer to our method as CSO (confidence-scaled U-OOD detector). The next sections describe our U-OOD detector~$\sigma^{in}$ and how it is used to define the few-shot learner~$s^-$.

\subsubsection{U-OOD detector:} 

Our U-OOD detector combines elements from the Mahalanobis anomaly detector (MahaAD)~\cite{rippel2021modeling}, known for its robustness and speed~\cite{doorenbos2022data}, and from non-parametric nearest-neighbor scoring methods~\cite{bergman2020deep,reiss2021panda,sun2022out}.
As in MahaAD, given the collection of ID~samples~$\D^\textrm{in} = \{\x_i\}_{i=1}^N$, we fit a Gaussian distribution parameterized by the data mean~$\bmu$ and covariance matrix~$\bm{\Sigma}$. To prevent numerical problems with near-singular covariance matrices in high-dimensional or low-data regimes, we use shrinkage following the standard hyperparameter-free method of~\cite{ledoit2004well}.
MahaAD uses the induced Mahalanobis distance~$d_{\covar'}(\x, \bmu)$ between~$\bmu$ and a test sample~$\x$ to estimate its OOD~score. We instead use this distance to perform a 2-NN search (following, e.g.,~\cite{bergman2020deep}) and score test samples with the average distance to their $2$~nearest neighbors,
\begin{equation}
    \sigmain(\x) = \dfrac{1}{2} \sum_{\x'\in N^2_{\covar'}(\x)} d_{\covar'}(\x, \x'),
    \label{eq:sigmain}
\end{equation}
where $N^2_{\covar'}(\x)$ denotes the 2-nearest neighbors of~$\x$ in the training data measured with the Mahalanobis distance induced by~$\covar'$. We refer to this as MkNN.

The detector~$\sigmain$ defined in Eq.~\eqref{eq:sigmain} is inappropriate for image samples, as the Mahalanobis distance is not a reliable measure of image similarity on high-dimensional spaces. Instead, when dealing with images, we first describe each image~$\x_i$ with a sequence of feature vectors~$\{\f_\ell(\x_i)\}_{\ell=1}^L$, where $\f_\ell$~denotes the result of applying global average pooling on the feature map of the $\ell$-th layer of a pre-trained convolutional neural network~$\f$. We then use the descriptors of the training images to build a collection of layer-wise detectors~$\{\sigmain_\ell\}_{\ell=1}^L$. In particular, the detector~$\sigmain_\ell$ at layer~$\ell$ applies the Eq.~\eqref{eq:sigmain} with the covariance matrix~$\covar'_\ell$ computed from the collection of features~$\{\f_\ell(\x_i)\}_{i=1}^N$. The final OOD score for a test image~$\x$ is the sum of the scores over all layers,
\begin{equation}
    \sigmain(\x) = \sum_{\ell=1}^L \sigmain_\ell(\f_\ell(\x)).
    \label{eq:sigmain2}
\end{equation}

\subsubsection{Few-shot learner $s^-$:}
Our few-shot learner extends the OOD detector~$\sigmain$ trained with~$\D^\textrm{in}$ by incorporating a twin detector~$\sigmaout$ trained with~$\D^\textrm{out}$. The OOD score of the few-shot learner is computed as the difference between both detectors,
\begin{equation}
    s^-(\x) = \sigmain(\x) - \lambda \sigmaout(\x),
\end{equation}
where the factor~$\lambda$ controls the influence of~$\sigmaout$ in the final score. The value of~$\lambda$ depends on the confidence levels of the detectors, which, in turn, rely on the contents in~$\D^\textrm{in}$ and~$\D^\textrm{out}$. 

To measure the confidence of a detector~$\sigma$ trained with the dataset~$\D$, we assess its variability under bootstrapping. More specifically, we produce $M$~bootstrap samples~$\{\D^{(m)}\}_{m=1}^M$ of $N$~elements randomly sampled from~$\D$ with replacement, and compute the covariance matrices~$\covar^{(m)}$ for each sample~$\D^{(m)}$. The uncertainty of the detector is measured as the variability of the bootstrapped covariance matrices, 
\begin{equation}
    U(\D) = \dfrac{1}{M d^2} \sum_{m=1}^M \left\| \covar^{(m)} - \bar{\covar} \right\|^2_F,
\end{equation}
where $\bar{\covar}= \frac{1}{M}\sum_m \covar^{(m)}$. The factor~$\lambda$ is then computed as the ratio between the uncertainties of the detectors,
\begin{equation}
    \lambda = \min\left(1, \gamma\dfrac{U(\D^\textrm{in})}{U(\D^\textrm{out})}\right),
\end{equation}
where $\gamma>0$~is a hyperparameter of our method. If no OOD~samples are available, $U(\D^{out})\to \infty$ and $\lambda = 0$, thus making the few-shot learner~$s^-$ equivalent to the base OOD detector~$\sigmain$.

When working with image data, we proceed layer by layer, as previously discussed for the U-OOD detector. In particular, we build a few-shot learner~$s^-_\ell$ per each layer~$\ell$ of the feature extractor~$\f$, and the final score is the sum of the layer-wise scores,
\begin{equation}
    s^-(\x) = \sum_{\ell=1}^L s^-_{\ell}(\f_\ell(\x)).
\end{equation}

%% file: 04experiments.tex
\section{Experiments}


\subsection{Experimental set-up}

\subsubsection{Datasets:} 
We introduce three IDE benchmarks to compare methods comprising various modalities, contamination ratios, and other settings. The samples for evaluating the methods and those in $\D_t^{\textrm{deploy}}$ do not overlap.
\begin{enumerate}
    \item NIH~\cite{tang2020automated}: Training: 4'261 healthy chest X-rays. Testing: 250 healthy scans as ID and 250 pathological chest X-rays as OOD. All methods see $T=10$ steps of $K=50$ test samples, with contamination $\pi=0.2$. 
    \item MURA~\cite{rajpurkar2017mura}: Training: 5'106 musculoskeletal radiographs of fingers. Testing: 250 finger scans as ID with scans of elbows, forearms, hands, humeri, shoulders, and wrists as OOD, where $T=5$, $K=100$, and $\pi=0.1$.
    \item DRD~\cite{graham2015kaggle}: Training: 25'809 healthy retinal fundus photographs as ID. Testing: 250 healthy scans as ID, with 250 scans of strongest level of diabetic retinopathy as OOD, where $T=5$, $K=50$, and $\pi=0.1$.
\end{enumerate}

\subsubsection{Baselines:}

We compare our method to seven baselines that comprise the top-performing methods from related fields: AdaODD~\cite{zhang2023model}, ETLT~\cite{fan2022simple}, SAL~\cite{du2024does}, BCE~\cite{liznerski2022exposing}, the Mahalanobis difference (MDiff)~\cite{sehwag2021ssd}, and HSC~\cite{liznerski2022exposing}. 
Furthermore, we include MahaAD~\cite{rippel2021modeling} as a U-OOD baseline. All methods use the same ImageNet pre-trained backbone. To ensure fairness, all methods use the same grid search procedure to set their hyperparameters. We measure the performance of each method on the CIFAR10 experiment~\texttt{Plane:Rest} and select the highest-performing configuration over the grid. 

\subsubsection{Evaluation metrics:} In contrast with previous works, we are interested in measuring the quality of the OOD detector over time. As such, we propose two metrics that consider this aspect: the \underline{A}rea \underline{U}nder the \underline{F}PR@95 curve (AUF) and the \underline{A}rea \underline{U}nder the \underline{A}UC curve (AUA). These are computed by first evaluating the FPR@95 and AUC at every timestep and plotting the resulting FPR@95/AUC curves with respect to $t$. Then, the AUF and AUA are given by the area under the FPR@95 and AUC curves, respectively, normalized by the time elapsed. 

\subsubsection{Implementation details:} 
For $s^-$, we extract features from $L=4$~layers at the end of every ResNet-18 block and normalize the features of the final layer following~\cite{fan2022simple,reiss2023mean}. The binary classifier $s^+$ also uses a ResNet-18 architecture. It is trained with Adam~\cite{kingma2014adam} using a learning rate of $10^{-5}$ and batch size~256 for ten epochs on NIH and an equivalent number of iterations on the other datasets. We apply data augmentation in the form of random resized crops, color jitter, and horizontal flips and initialize the binary classifier with the weights from the previous timestep. At every step, the elements of $\D_t^\textrm{deploy}$ are re-labeled with the current model $s_t(\x)$ using a threshold. The threshold is determined on the training data such that 95\% is considered in-distribution. We standardize the scores of $s^+$ and $s^-$ before combining them to ensure they have similar scales. The hyperparameters $\beta$ and $\gamma$ were set via the grid search to $1/300$ and $3$, respectively. We found CSO robust to their settings, as shown in the next section.

\begin{table}[t]
\centering
\caption{\textbf{Comparative evaluation.} We report the mean of the AUF ($\downarrow$) and AUA ($\uparrow$) over five trials. \textbf{Bold} and \underline{underlined} indicate best and second best, respectively. Our method obtains the best performance overall. }
\begin{tabular}{lcccccccc}
\toprule
& \multicolumn{2}{c}{NIH} & \multicolumn{2}{c}{MURA} & \multicolumn{2}{c}{DRD}  & \multicolumn{2}{c}{Mean} \\
\cmidrule(lr){2-3} \cmidrule(lr){4-5} \cmidrule(lr){6-7} \cmidrule(lr){8-9} 
& AUF & AUA & AUF & AUA & AUF & AUA & AUF & AUA \\
\midrule
SAL & 91.5\tiny{$\pm$6.3} & 58.3\tiny{$\pm$12.1} & 90.0\tiny{$\pm$8.3} & 56.6\tiny{$\pm$9.2} & 77.4\tiny{$\pm$8.9} & 67.3\tiny{$\pm$7.9} & 86.3 & 60.7\\
HSC & 70.4\tiny{$\pm$0.9} & 78.6\tiny{$\pm$0.2} & 68.8\tiny{$\pm$6.5} & 79.8\tiny{$\pm$2.3} & 81.8\tiny{$\pm$2.0} & 71.6\tiny{$\pm$1.3} & 73.7 & 76.7 \\
ETLT & 54.9\tiny{$\pm$4.0} & 82.0\tiny{$\pm$1.0} & 61.5\tiny{$\pm$4.9} & 86.5\tiny{$\pm$0.7} & 64.8\tiny{$\pm$2.0} & 78.3\tiny{$\pm$1.1} & 60.4 & 82.3 \\
BCE & 60.1\tiny{$\pm$16.7} & 74.0\tiny{$\pm$9.6} & 56.8\tiny{$\pm$8.2} & 84.3\tiny{$\pm$3.1} & 60.5\tiny{$\pm$16.9} & 80.3\tiny{$\pm$7.3} & 59.1 & 79.5 \\
MahaAD & 53.8\tiny{$\pm$3.9} & 85.3\tiny{$\pm$0.9} & 47.0\tiny{$\pm$4.5} & 90.8\tiny{$\pm$0.7} & 71.4\tiny{$\pm$3.9} & 78.1\tiny{$\pm$1.0} & 57.4 & 84.7 \\
AdaODD & \underline{46.9}\tiny{$\pm$5.5} & \underline{87.0}\tiny{$\pm$0.9} & 48.3\tiny{$\pm$7.9} & 88.6\tiny{$\pm$1.3} & \textbf{57.8}\tiny{$\pm$4.4} & \textbf{83.5}\tiny{$\pm$1.1} & 51.0 & \underline{86.4} \\
MDiff & 48.1\tiny{$\pm$4.4} & 83.0\tiny{$\pm$2.1} & \underline{42.9}\tiny{$\pm$6.6} & \underline{91.3}\tiny{$\pm$1.1} & \underline{60.3}\tiny{$\pm$4.6} & 82.7\tiny{$\pm$3.1} & \underline{50.4} & 85.7\\
\rowcolor{mygray} CSO (ours) & \textbf{34.9}\tiny{$\pm$6.1} & \textbf{90.9}\tiny{$\pm$1.7} & \textbf{37.5}\tiny{$\pm$5.6} & \textbf{91.8}\tiny{$\pm$0.9} & 61.9\tiny{$\pm$5.6} & \underline{83.1}\tiny{$\pm$1.6} & \textbf{44.8} & \textbf{88.6} \\
\bottomrule
\end{tabular}
\label{tab:highres}
\end{table}

\subsection{Results}
\cref{tab:highres} reports the results on our benchmark. Of all methods, SAL, HSC, ETLT, and BCE do not outperform the unsupervised MahaAD baseline on average. 
BCE is especially inconsistent; for instance, it achieves excellent results on DRD but performs below average in other cases, demonstrating the need for a more robust method.
In contrast, AdaODD and MDiff score consistently high in all experiments. Nonetheless, they are outclassed by CSO, which reaches the best score for NIH and MURA and the best overall performance by 5.6 AUF and 2.2 AUA compared to the next-best method.

The performance evolution over time for some of the best methods on NIH is shown in \cref{fig:curves}(a). All methods benefit from incorporating unlabeled samples, with our method improving fastest. The only exception is AdaODD, whose optimal hyperparameters from the grid search assign low importance to the unlabeled NIH data.


%% file: 05discussion.tex
\subsection{Ablations}

\subsubsection{Hyperparameter sensitivity}

From the grid search conducted on \texttt{Plane:Rest}, we find CSO to be robust to its main hyperparameters $\beta$ and $\gamma$: all combinations with $\gamma\in[1,3,5]$ and $\beta\in[1/100,1/300,1/500]$ achieve between $92.6$ and $93.2$ AUA. 

\subsubsection{Contamination ratio}

\begin{figure}[t]
  \centering
  \setlength\tabcolsep{10pt}
  \begin{tabular}{cc}
    \includegraphics[width=0.35\linewidth]{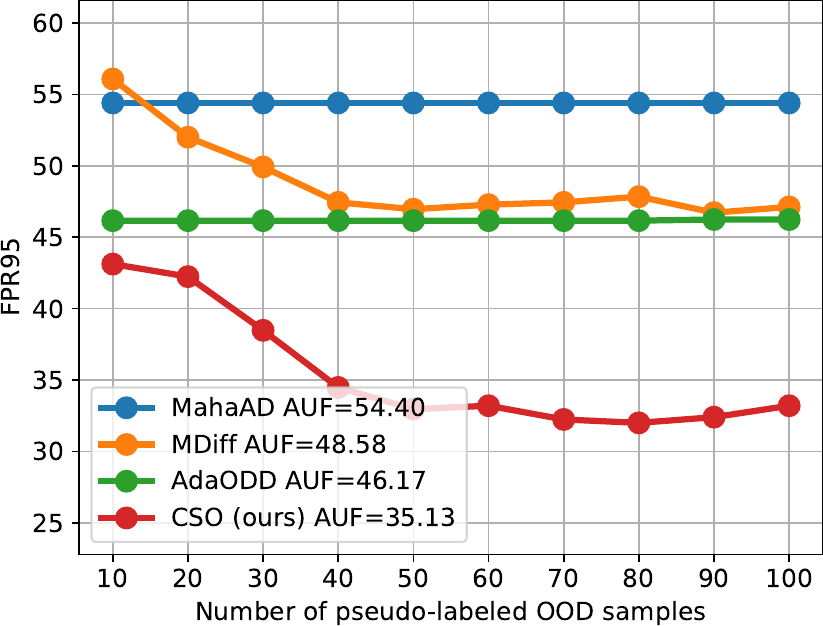} &
    \includegraphics[width=0.35\linewidth]{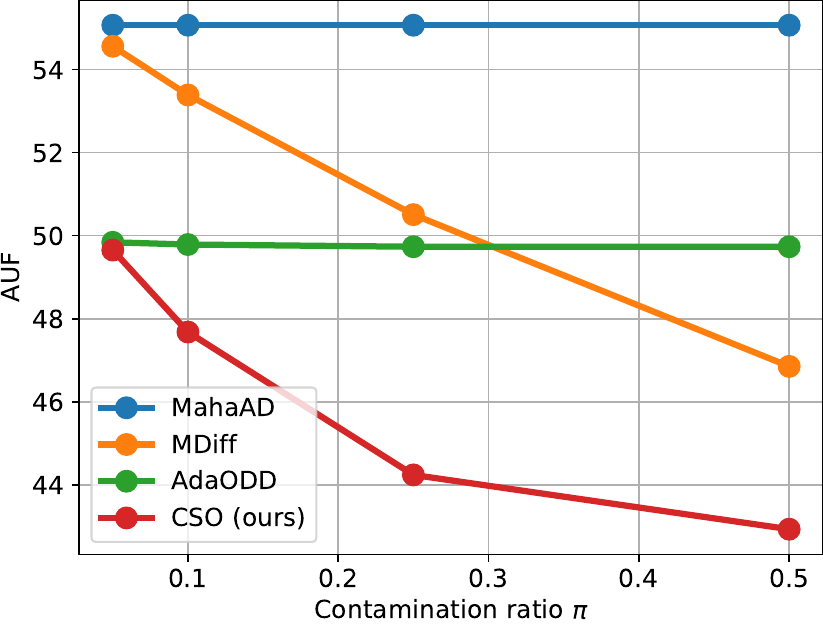} \\
    (a) & (b)\\
  \end{tabular}
    \caption{\textbf{Ablation curves.} In (a), we show how the FPR@95 evolves over time for the best methods on NIH. Our method achieves the best results, already after one iteration. In (b), we compare methods by AUF under varying contamination ratios on NIH. Our method consistently outperforms the baselines at different contamination levels.}
    \label{fig:curves}
\end{figure}

We probe the effectiveness of CSO compared to the three best baselines under varying contamination ratios $\pi$ while keeping the number of OOD samples per step fixed at ten in \cref{fig:curves}(b). 
As expected, all iterative methods benefit from having a higher fraction of unlabeled OOD samples in the test set. 
Nonetheless, CSO achieves the best AUF for all contamination levels.

\subsubsection{Scoring function}

We ablate our design choices by showing that (1) MkNN outperforms both kNN and MahaAD for U-OOD detection and (2) $s^-$ outperforms unscaled scoring functions on few-shot OOD. To do so, we run experiments with our method on the standard one-class CIFAR10 benchmark. 

From \cref{tab:abl}(left), MkNN outperforms the kNN scoring by 5.9 AUC and MahaAD by 1.6 AUC, showcasing its practical usefulness over top-performing U-OOD detectors~\cite{doorenbos2022data}. \cref{tab:abl}(right) shows that equipping MDiff with our confidence scaling, which we label as $s^-_{Maha}$, already improves few-shot results. We improve the results by a further 0.6 and 0.8 AUC using MkNN. These results on natural images also confirm the usefulness of our method beyond medical settings.

\begin{table}[t]
\begin{center}
\caption{\textbf{Ablation study on CIFAR10.} (left) U-OOD performance in AUC over one run as the methods are deterministic. MkNN outperforms the other U-OOD scoring functions. (right) Comparing few-shot OOD performance. We report the mean AUC over five runs. $n$-shot refers to using $n$ ground-truth OOD samples. The confidence scaling is important to achieve the best results. }
 \begin{tabular}{lc}
\toprule
 & AUC\\
\midrule
kNN & 81.7 \\
MahaAD & 86.0 \\
\rowcolor{mygray} MkNN (ours) & \textbf{87.6} \\
\bottomrule
\end{tabular}
\quad\quad
 \begin{tabular}{lcc}
\toprule%
 & \textit{5-shot AUC} & \textit{10-shot AUC} \\
 \midrule
kNN & 75.6 & 79.8\\
MDiff & 84.7 & 89.9\\
\rowcolor{mygray} $s^-_{Maha}$ (ours) &  91.2 & 91.9 \\
\rowcolor{mygray} $s^-_{MkNN}$ (ours) & \textbf{91.8} & \textbf{92.7}\\
\bottomrule
\end{tabular}
\label{tab:abl}
\end{center}
\end{table}

%% file: 06conclusion.tex
\section{Conclusion}

We introduced the setting of IDE for U-OOD detection, which reflects the iterative process of real-world model deployment, along with new metrics and benchmarks for the task. Furthermore, we presented CSO, a novel method for IDE that gradually transforms the base U-OOD detector into a binary classifier. In doing so, we additionally introduced an OOD scoring function that uses the Mahalanobis distance to compute a nearest-neighbors score, and a few-shot OOD detector that takes into account the confidence of the distributions involved. Extensive experiments showed that our simple approach outperforms methods from several related fields.